\documentclass[pdflatex,sn-mathphys]{sn-jnl}

\usepackage{amsmath}

\jyear{2021}%

\theoremstyle{thmstyleone}%
\usepackage[center]{caption}
\theoremstyle{thmstyletwo}%

\theoremstyle{thmstylethree}%

\begin{document}

\title[IRTCI: Item Response Theory for Categorical Imputation]{IRTCI: Item Response Theory for Categorical Imputation}

\author*[1]{\fnm{Adrienne} \sur{Kline}}\email{askline1@gmail.com}
\author*[1]{\fnm{Yuan} \sur{Luo}}\email{yuan.luo@northwestern.edu}

\affil[1]{\orgdiv{Department of Preventative Medicine}, \orgname{Northwestern University}, \orgaddress{\street{Lake Shore Dr}, \city{Chicago}, \postcode{60611}, \state{IL}, \country{USA}}}

\abstract{Most datasets suffer from partial or complete missing values, which has downstream limitations on the available models on which to test the data and on any statistical inferences that can be made from the data. Several imputation techniques have been designed to replace missing data with stand in values. The various approaches have implications for calculating clinical scores, model building and model testing. The work showcased here offers a novel means for categorical imputation based on item response theory (IRT) and compares it against several methodologies currently used in the machine learning field including k-nearest neighbors (kNN), multiple imputed chained equations (MICE) and Amazon Web Services (AWS) deep learning method, Datawig. Analyses comparing these techniques were performed on three different datasets that represented ordinal, nominal and binary categories. The data were modified so that they also varied on both the proportion of data missing and the systematization of the missing data. Two different assessments of performance were conducted: accuracy in reproducing the missing values, and predictive performance using the imputed data. Results demonstrated that the new method, Item Response Theory for Categorical Imputation (IRTCI), fared quite well compared to currently used methods, outperforming several of them in many conditions. Given the theoretical basis for the new approach, and the unique generation of probabilistic terms for determining category belonging for missing cells, IRTCI offers a viable alternative to current approaches.}




\maketitle

\section{Introduction}\label{sec1}
The purpose of this investigation was to introduce a new approach to imputing missing data for categorical variables – Item Response Theory for Categorical Imputation (IRTCI). Imputing missing values for categorical data has proven problematic, much more so than for continuous, normally distributed data \cite{finch2010imputation}. When data include large numbers of categorical data, multiple imputation techniques are challenging, as the space of potential models is enormous \cite{akande2017empirical}. Several attempts to deal with this problem have been introduced, including multinomial and log-linear models \cite{schafer1999multiple}, clustering \cite{dinh2021clustering}, \cite{vidotto2015multiple} and a variety of multiple imputation methods such as expectation-maximization with bootstrapping, correspondence, latent class analysis, hot deck, and chained equations \cite{stavseth2019handling}. Borrowing from psychometric theory, Item Response Theory (IRT) offers a family of models that have been designed specifically to handle categorical data. The process results in a series of probabilities to determine whether the missing value belongs to a particular category. Demonstrating how to leverage these models for use in imputing missing data is the purpose of the current study.\\

\subsection{Missing Data}
Many datasets suffer from being incomplete, in that they have missing data points in some or all variables. Missing data can occur for many reasons including, but not limited to: hardware limitations (i.e. sensor drop out), subject loss at follow-up (e.g. patient who did not return or dies), data entry errors, rare events, non-response (i.e. surveys), or the data were intentionally not collected for a case specific reason. How to best handle missing data can be difficult to resolve, especially when the causal reason for it remains unknown. Some statistical procedures cannot function with missing values and automatically eliminate cases with missing data, such as factor analysis, Cronbach’s alpha and many feed forward neural networks. Even if only a few data points are missing from each variable, the effect of dropout, if performed case-wise, may result in a reduction of power of the statistical test, not having enough data to perform the analysis, or misleading findings if the remaining cohort is not a random sample of all cases. Similarly, many machine learning (ML) models cannot handle missing values, such as support vector machines, GLMnet, and neural networks. The few models that are able to tolerate missing values are Naive Bayes and some tree based models under the CART methodology \cite{Breiman1984}.\\

Missing data can be classified into three categories \cite{dong2013principled}; missing completely at random (MCAR), missing at random (MAR) and missing not at random (MNAR). MCAR data follows the logic that the probability of an observation being missing does not depend on observed or unobserved measurements. MAR missing data is conditional on one or more covariates, where the probability of an observation being missing depends only on observed variables in the dataset. Because of the characteristics of MCAR and MAR data, they are amenable to data-driven approaches to handling them. However, when observations are neither MCAR nor MAR, they are classified as MNAR, meaning the probability of an observation being missing depends on unobserved variables/information not available in the analysis. The missing mechanism, then, needs to be theoretically justified and incorporated into the data. Because of the ‘top-down’ nature of handling MNAR data, this type of missing date will not be discussed in the current study. \\

\subsection{Traditional Imputation Techniques}
When more than 40\% of data from important variables are missing, it is recommended that any inferences drawn should be exploratory; conversely, with less than 5\% of the data missing dropping cases or simple scalar substitutions are warranted \cite{madley2019proportion}. However, if only smaller portions of data are missing, preserving as much of the information becomes important, particularly with smaller data sets, leading to the need of imputing values to substitute into the missing cells. Several imputation techniques have been developed to do so. Some common examples are forward fill, backward fill, mean or most frequent, and Bidirectional Recurrent Imputation for Time Series (BRITS) \cite{CaoBRITS}. Forward and backward fill work by carrying the most recent value forward or backward, respectively, filling in where appropriate. Imputing with the mean, median or mode works by computing the value of the mean, median or mode in relation to the column and filling this value in where missing. BRITS substitution is specific to time series data.\\

\subsection{Item Response Theory (IRT)}
The concept of IRT for imputation was introduced by \cite{huisman2001imputation}. However, this does not perform a comparison with current state-of-the-art (SOTA) methods, how this impacts downstream predictive tasks and the various algorithmic adaptations for ordinal, nominal and binary imputation. The purpose of this study is to demonstrate how this technique can be used for imputation and compare its effectiveness with three of the more traditional imputation methods and how this has impacts downstream machine learning tasks. IRT is a family of mathematical models that link underlying unobserved (latent) traits of individuals/cases to the pattern of their responses to a series of observed variables (i.e., items or features) \cite{embretson2013item}. This linkage is manifested as one or more logistic functions that specify the probability of obtaining a specific value on any feature as a function of a case’s underlying trait value. These logistic functions are generated using a maximum likelihood iterative approach that analyzes the entire pattern of all feature values for all cases simultaneously. IRT assumes that the latent trait is organized along a continuum called theta ($\theta$) and all individual cases are placed along that continuum. Higher values of $\theta$ are associated with higher levels of the underlying trait. It is assumed that higher values on the features are also associated with higher values of $\theta$.\\
As part of the analysis process, characteristics of the features, such as their difficulty and discrimination, are estimated as well as an estimate of each case’s standing along the underlying trait – their theta ($\theta$) score. Because IRT is an individual measurement theory, it was developed \cite{Lord1968} \cite{Rasch1960} and is currently used primarily in the psychological and educational literatures to assess the psychometric properties of items and tests. However, IRT has been used in the machine learning literature to assess the utility of features \cite{Kline2021}, natural language processing systems \cite{LalorJohnP2016BaES}; and classifiers \cite{Kline2020}; \cite{Martinez-PlumedFernando2019Irti}.

The current study assesses how well IRT performs as a mechanism for imputation of missing feature data. IRT focuses on the pattern of all the available observed feature values to generate each case’s overall $\theta$ score. Then the imputed missing values are based on each individual case’s $\theta$ score. Because IRT uses all the feature information available for all cases, it is possible to impute valid values for those cases with missing data. One important result, then, of IRT imputed values is that they do not incorporate the outcome variable values in the protocol, as do many other imputation methods. In doing so, IRT avoids the circularity of using the classification outcome to impute missing values. This avoids the problem of overly optimistic findings in predictive modeling studies, when using the outcome to set values for a predictor that is then used to predict that same outcome. Such outcome information would not be available to classify/predict prospective new cases.

Three members of the family of IRT models will be used in the current study. One is the 2-parameter logistic model (2-PL) \cite{EmbretsonSusanE2000Irtf} used when the features are coded in a binary (0, 1) way. Another is the Graded Response Model (GRM) \cite{SamejimaFumiko1970EEol} used when features have ordinal-level values. Since IRT analyses do not handle continuous interval level data, such data can be converted into multiple ordinal level categories and run using the GRM. The third IRT model is the Nominal Response Model (NRM) used when feature values are nominal/categorical \cite{DarrellBockR.1972Eipa}. Salient attributes of these imputation methods are listed in Table 1.\\

\begin{table}[!h]
 \caption{Imputation Types and Attribute Comparison}
  \centering
  \begin{tabular}{lllll}
  \toprule
	Attribute & \multicolumn{3}{c}{Imputation Method}\\
	\cmidrule(l){2-5}
	 & KNN & MICE & Datawig & IRT\\ 
	\midrule
 	Categorical imputation  &  \checkmark$^a$ & \checkmark$^a$ & \checkmark & \checkmark\\\
  	Scalable  &   x &  \checkmark$^b$ &  \checkmark & \checkmark\\
  	Uses outcome  &  x  & x &   \checkmark & x \\
  	Works for time series  &  x$^c$  & \checkmark &  \checkmark &  \checkmark \\
  	Works for small datasets & \checkmark  & \checkmark &  x$^d$ &  \checkmark \\
    \bottomrule
  \end{tabular}
  \footnotetext[a]{both KNN and MICE require categorical to be ordinal or be transformed into one-hot encoded if nominal}
  \footnotetext[b]{Depending on the length of the dataset}
  \footnotetext[c]{Theoretically possible, however, computationally difficult}
  \footnotetext[d]{Theoretically possible, however, likely unreliable}
  \label{tab:table}
\end{table}

\section{Methods}
\subsection{Datasets}
Three different data sets were selected for this study: Diamonds \cite{Agrawal2017}; Housing \cite{RUBENS2020} and Heart Disease \cite{TEBOUL2015}. These were selected because they: 1) use different types of categorical data to be imputed (ordinal, binary and nominal), 2) have an outcome to allow for predictive utility assessment, and 3) are complete (no missing values), so the ground truth for the missing cases were available to compare different imputation methods. Thus, they provided a broad comparative field regarding how IRT performs relative to other imputation methods.\\
Within each data set, a single predictor variable was selected to be missing. Null values substituted in each of these specified predictor variables in four different amounts (missing 5, 10, 30 and 50\%), each following two different structures (MCAR vs MAR). Therefore, each dataset gave rise to eight unique datasets for imputation. To generate the MCAR type data sets, values were randomly replaced with null values. Generating MAR data was performed on a per dataset instance by first identifying a conditional variable on which to generate the MAR data sets. The files were then sorted on the conditional variable and 5, 10, 30 and 50\% of the target missing variable was removed from the top of the dataset. To verify MCAR versus MAR missing data structures, Little’s test was used \cite{Li2013}. Little’s test is a modified chi-Square test to determine if one or more systematic relationships between the missing data and other variables exist and is expected to be significant in MAR data sets and non-significant in MCAR data sets.\\

One issue that arose was that since IRT does not accommodate continuous data, such features had to be re-coded into ordinal-level categories, as this is required for use in the GRM analyses. To do so, histograms of the data were generated for each continuous feature and cut points made to preserve the original shape of the distribution, as many of the feature variables were non-normally distributed. Data that were affected in such a way were split into quartiles, providing four-level ordinal variables. This conversion was only done when running the IRT imputations. \\

\subsubsection{Ordinal Imputation Data}
The diamonds data is a set of 53,920 diamond cases with a continuous outcome (price). The eight features are a combination of ordinal (e.g., clarity) and continuous (e.g., dimensions along x, y, z). The feature that was selected to be missing for purposes of this study was color (an ordinal variable with 8 different levels). Other variables included price in US dollars (\$326-\$18,823), carat weight of the diamond (0.2-5.01), cut quality of the cut (Fair, Good, Very Good, Premium, Ideal), color; from J (worst) to D (best), clarity; (I1 (worst), SI2, SI1, VS2, VS1, VVS2, VVS1, IF (best)), x length in mm (0-10.74), y width in mm (0-58.9), z depth in mm (0-31.8), depth total depth percentage = z / mean(x, y) = 2 * z / (x + y) (43-79) and table width of top of diamond relative to the widest point (43-95). A list of the variables and their codes are shown in Table 2. The conditional variable to generate the MAR data sets was ‘carat size’ in this data set.\\

\begin{table}[h]
\centering
 \caption{Diamond dataset}
  \begin{tabular}{ll}
    \toprule
    Feature  &   Feature Type\\
    \midrule
    Carat &numeric - continuous \\
    Cut   &  continuous   \\
    Color     &  ordinal  \\
    Depth     &  numeric - continuous  \\
    X  (mm)   &  numeric - continuous  \\
    Y   (mm)  &  numeric - continuous \\
    Z   (mm) &  numeric - continuous \\
    Price (outcome)  &  numeric - continuous \\
    \bottomrule
  \end{tabular}
  \label{tab:table}
\end{table}

\subsubsection{Binary Imputation Data}
The heart disease data is a set of 253,680 responses from Behavioral Risk Factor Surveillance System (BRFSS) 2015, generated by the CDC to be used for the binary classification of heart disease/attack (no heart disease – coded 0, heart disease – coded 1). 23,893 of the cases had heart disease. An equivalent number were randomly selected from the non-heart disease cases, producing a final and balanced data set of 47,786 cases. A list of the variables and their codes are shown in Table 3. The feature that was selected to be missing for purposes of this study was high blood pressure (a binary variable). The conditional variable to generate the MAR data sets was ‘age’ in this data set.

\begin{table}[h]
 \caption{Heart Disease Dataset}
  \centering
  \begin{tabular}{ll}
    \toprule
    Feature  & Feature Type\\
    \midrule
    BMI &   numeric - continuous\\
    Age   &  numeric - continuous\\
    Smoker    &   0,1 - binary \\
    Stroke & 0,1 - binary\\
    Diabetes & 0,1 - binary\\
    No Physical activity & 0,1 - binary\\
    No Vegetables & 0,1 - binary\\
    Difficulty Walking  &  0,1 - binary  \\
    High Cholesterol    &  0,1 - binary \\
    High Blood pressure    &  0,1 - binary\\
    Heart Disease or Attack (outcome)  &  0,1 - binary\\
    \bottomrule
  \end{tabular}
  \label{tab:table}
\end{table}

\subsubsection{Nominal Imputation Data}
The housing data set is made up of 10,692 unique rental units and their features with a continuous outcome (rental price). Other features included whether the space was furnished or not, number of rooms, square footage, number of bathrooms and the city in which it was located. The feature that was selected to be missing for purposes of this study was city (a nominal categorical variable with 5 unique values). The conditional variable to generate the MAR data sets was ’number of rooms’ in this data set.

\begin{table}[h]
 \caption{Housing Dataset}
  \centering
  \begin{tabular}{lll}
    \toprule
    Feature  & Feature Type\\
    \midrule
    City & categorical - nominal\\
    Area   & numeric - continuous  \\
    Rooms     &  numeric - continuous \\
    Bathrooms   &  numeric - continuous  \\
    Furniture    &  numeric - binary \\
    Rent price (outcome)  &  numeric - continuous \\
    \bottomrule
  \end{tabular}
  \label{tab:table}
\end{table}

\subsection{Imputation Methods}
\subsubsection{Existing methods}
Three commonly used, robust imputation methods were employed in this study, k-NN, MICE, and a deep learning method called DataWig. K-NN works very much like the algorithm for classification. The substituted value is based on a specified number ’k’ of the closest point estimates in an n-dimensional space. MICE also known as Sequential Regression Imputation, was developed by Rubin \cite{Rubin1987} and leverages a series (chain) of regression equations to obtain imputation values. MICE starts with a simple imputation method such as mean substitution. However, the process is repeated several times on different portions of the data and regressed on other variables, where the final imputed value is one that converges to stability. DataWig is a deep learning imputation method developed by Amazon Web Services (AWS) \cite{datawig} that uses a Long Short Term Memory network (LSTM). It follows a similar approach as that of MICE that can be extended to allow for different types of data (categorical, numerical, text) to be used when imputing missing values. For categorical variable imputation, an EmbeddingFeaturizer is used, where training data are comprised of rows of complete data and the training supplies the remaining structured dataset. The predicted outcome is the value to be imputed and is subsequently substituted into the final dataset. \\

\subsubsection{IRT Imputation}
IRT provides an alternative approach to imputation, as described earlier. The IRTPRO (Vector Psychometric Group, 2021) program was used to estimate the IRT feature and case parameters in all data sets. Data can be imported into the program from a number of different file formats, including the type used in this study (.csv). All missing data cells were coded with -1. The interface allows for a mixture of different types of features within the same analysis (i.e., a mix of binary, ordinal, or categorical features can be used in the same analysis). Each model was specified to be based on one group of cases using a unidimensional set of features. IRTPRO uses maximum likelihood to estimate feature parameters and expected a posteriori (EAP) to generate a $\theta$ score for each case. Parameters are estimated in the logistic metric. Some programs have historically rescaled the parameters to approximate the normal to give function, but this is not done in IRTPRO as has been suggested more recently \cite{CamilliGregory1994OotS}. Each feature was specified to follow a 2-PL, GRM or NRM. \\
If the response by examinee $j$ to item $i$ is denoted by a random variable $U_{ij}$, it is convenient to code the two possible scores as $U_{ij}$ = 1 (correct) and $U_{ij}$ = 0 (incorrect). To model the distribution of this variable, or, equivalently, the probability of a correct response, the ability of the examinee is presented by a parameter $\theta$ $\in$ (-$\infty$,+$\infty$), and it is assumed in a two-parameter model that the properties of item i that have an effect on the probability of a success are its difficulty and discriminating power, which are represented by parameters $b_i$ $\in$ (-$\infty$,+$\infty$) and $a_i$ $\in$ (0,+$\infty$), respectively. The probability of success on a given item $i$ is usually represented s $P_i(\theta)$. The 2-PL generates a threshold ($b$) and slope ($a$) for each feature and a $\theta$ for each case. Using these estimated parameters, the linking function between the underlying trait and particular feature can be described as follows: 

\begin{equation}
    Pij(U_{ij}=1\mid\theta) = \frac{e^{a_{i}(\theta - b_{ij})}}{1+ e^{a_{i}(\theta - b_{ij})}}
\end{equation}

\medskip

Using the representation in Eq. 1, the joint likelihood function estimation for simultaneously computing ability $\theta$ and parameters $a_i$ and $b_i$ in the case of N examinees and n items associated with the 2-PL model can be written as:

\begin{equation}
    L(\theta, a,b;u) = \prod_i\prod_j P_i(\theta_j;a_i,b_i)^{u_{ij}}[1-P_i(\theta_j;a_i,b_i)]^{1-u_{ij}}
\end{equation}

\medskip
Where $\theta\equiv(\theta_1$,...,$\theta_n$), $a\equiv(a_1,...,a_n$), $b\equiv(b_1,...,b_n$), $u\equiv(u_{ij})$, and $u_i$ and $u_j$ are the marginal sums of the data matrix, based on the response pattern $x$ across items (variables) and across an examinee (row case) \cite{lord1968statistical}. Maximizing the logarithm of the likelihood function results in the following set of estimation equations:

\begin{equation}
\begin{split}
\sum_i(a_i(u_{ij}-P_i(\theta_j;a_i,b_i))=0, j=1,...,N\\
\sum_j(a_i(u_{ij}-P_i(\theta_j;a_i,b_i))=0,  i=1,...,n\\\
\sum_j(u_{ij} - P_i(\theta_j;a_i,b_i))(\theta_j-b_i), j=1,...,n\\\
\end{split}
\end{equation}
The binary model in equation 1 has the simple interpretation of the probability of success being equal to the value of the person parameter $\theta$ relative to the value of the item parameters. The probability of being in the “1” category on a particular item $i$ can be ascertained for any case with a specific $\theta$-value. Using this model, a missing binary variable can be imputed – cases with probabilities below 50\% are imputed as 0 and those with probabilities above 50\% are imputed as 1. Figure 1a) showcases the curve for this model, where ability $\theta$ is a row/case characteristic and parameter values are associated with the variable (item). 

\begin{figure}[h!]
    \centering
    \includegraphics[scale=0.35]{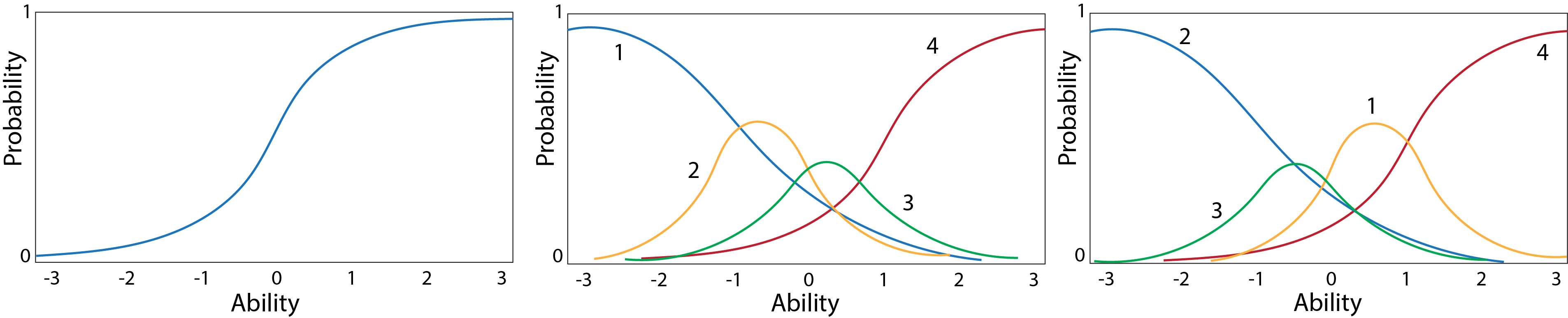}
    \caption{a) 2 Parameter Logistic Model b) Graded response model c) Nominal response model}
    \label{fig:my_label}
\end{figure}
The graded response model (GRM) represents a family of mathematical models that deals with ordered polytonomous categories, seen in Figure 1B. It uses a two-step process in to link the trait to features \cite{SamejimaFumiko1970EEol}. In the first step, a series of 2-PL functions for each of the category option boundaries are generated. For example, if one has a 5-option feature (coded 0, 1, 2, 3, 4), there would be 4 boundary functions: above 0 but less than 1, above 1 but less than 2, above 2 but less than 3 and above 3 but less than 4. In this first step, threshold parameters for each of the features’ option boundaries and an overall slope parameter for the feature are generated. If $\theta$ is the latent ability, and $U_i$ is a random variable to denote the graded item response to item i, and $u_i=(0,1, ... ,m_i)$ denotes the actual responses. The category response function, $P_{ui}(\theta)$, is the probability with which an examinee with ability $\theta$ receives a score $u_i$ is:

\begin{equation}
    P_{ui}(\theta)\equiv P[U_i=u_i \mid \theta]
\end{equation}

\medskip
Probabilities based on the other combinations, given $\theta$, are computed by subtracting the adjacent $P^*_{ik}(\theta)$:

\begin{equation}
    P^*_{ik}(\theta) = P^*_{ik}(\theta) - P^*_{ik+1}(\theta)
\end{equation}
\medskip

Therefore, in expanding Eq. 5 for a 5 category GRM, we would get:

\begin{equation} \label{eq1}
\begin{split}
    Option 0: P_{i0}(\theta) & = 1.0 - P_{i1}(\theta)\\
    Option 1: P_{i1}(\theta)   & = P_{i1}(\theta) - P_{i2}(\theta)\\
    Option 2: P_{i2}(\theta) &  = P_{i2}(\theta) - P_{i3}(\theta)\\
    Option 3: P_{i3}(\theta)& = P_{i3}(\theta) - P_{i4}(\theta)\\
    Option 4: P_{i4}(\theta) & = P_{i4}(\theta) - 0.0
\end{split}
\end{equation}

And the marginal likelihood solution (used for GRM) in can be written as:

\begin{equation}
    L(\theta, a,b;u)=\prod^N_{j=1}P_{x_j}
\end{equation}

Where $x_j$ is the response pattern obtained by an examinee $j$, $P_{xj}(\theta)$ for an examinee $i$ equals the joint likelihood function, $L(\theta_j, a_i, b_{ui})$. In this equation, the probability of being coded “1” on a particular category $j$ of a feature $i$ can be ascertained for any case with a specific $\theta$-value. These functions generate probabilities associated with $m$ dichotomies. Continuing with the example of 5 categories the dichotomies would refer to the probability of being coded 1: 1) in category 0 contrasted with categories 1, 2, 3, and 4; 2) in categories 0, 1 contrasted with categories 2, 3, and 4; 3) in categories 0, 1, 2 contrasted with categories 3 and 4; 4) in categories 0, 1, 2, 3 contrasted with category 4. The second step of the process uses subtraction between the probabilities for each option boundary of that feature to estimate the probabilities for each option. The probability of responding at the lowest option or above is 1.0 and the probability of responding above the highest alternate is 0.0. The option probabilities are generated for each alternative in the 5-point scale as above: Using this model, missing ordinal cells are imputed and categories assigned for each case based on the category with the highest probability.\\

The nominal response model (NRM) also uses a two-step process (divide-by-total) to link the ability with features \cite{darrell1972estimating}. In a typical nominal response model, a person $N$ responds to each of $n$ items, where the item $i$ admits responses in $m_i$ mutually exclusive categories, as in the case with a multiple choice exam, seen in Figure 1C. In the first step, functions for each of the category options are generated by estimating the slopes $a$ and intercepts $c$ for each option. Based on a case’s score, the probability of being coded “1” on a particular category $j$ of a feature $i$ is calculated as the ratio of the probability of being in that category divided by the sum of the probabilities of falling into any of the categories on that feature, see Eq. 8.

\begin{equation}
    P_{ij}(\theta) = \frac{\exp(a_{ij}\theta + c_{ij})}{\sum_{x=0}^m \exp(a_{ij}\theta + c_{ij})}
\end{equation}
\medskip

The probability of the response pattern $U_iv= [U_{1\ell}, U_{2\ell},...,U_{n\ell}]$ as a function of $\theta$ can be represented by:

\begin{equation}
    L_{\ell}(\theta) = P(U_{\ell}\mid\theta)= \prod P_{ih}(\theta)
\end{equation}
\medskip

where $h=U_{i\ell}$ is the item score designating the category to which the response to item $i$ in pattern $\ell$ corresponds. $L_{\ell}$ is called the likelihood of $\theta$, given the pattern $Ue$, and $P_{\ell}$ is called the marginal probability of $U_{\ell}$. To ensure model identification in NRM, one of two constraints must be set for parameter estimation. Either the sum across feature slopes and feature intercepts must be set to zero ($\theta a_{ij} = \theta c_{ij} = 0$), or the lowest response category for each feature must be set to zero ($a_{i1} = c_{i1} = 0$). The IRTPRO program opts for the latter of these two constraint options, as has been suggested to be more plausible \cite{ThissenDavidARMf}. As with the GRM, this analysis estimates the category into which the case is most likely to fall. For imputation, each category is calculated based on parameters $a_i$ and $c_i$ and ability $\theta$, and the category with the highest probability becomes the corresponding imputed value for nominal level data.\\

\subsubsection{Imputation Assessments}
Factorial Analyses of Variance (4-levels of methodology * 2-levels of missing type) were conducted on the assessments. Follow-up tests using a Bonferroni correction were used to assess any differences between imputation methodologies and MAR and MCAR findings. Only significant results are reported. To assess the imputations relative to the complete datasets, F1 scores (Eq. 10) were calculated for the cells that had been imputed.

\begin{equation}
    F1 = 2*\frac{precision*recall}{precision + recall}
\end{equation}
\medskip

Additionally, machine learning models were trained to compare relative predictive utility between the different imputation methods and with the original complete data (ground truth). Several machine learning methods were trialed on the original data sets which included Linear Regression, Bayesian Ridge regression, Random Forest Regressor and XGBoostRegressor for the regression outcome data sets (Diamonds and Housing). Random Forest, neural network (NN), support vector machine (SVM) and XGBoost algorithms were used for classification outcome (Heart Disease Data set). Hyperparameters were determined using a random search within the various algorithms. The best model for each dataset was determined using the original dataset and then used with the imputed datasets to allow for a consistent comparison.\\
Root Mean Square Error (RMSE) summary values for the Diamond and Housing outcome predictions were used to assess fit of the expected to observed values, where lower values are better. Area Under the Curve (AUC) was used to assess the models’ capability of distinguishing between classifications for the Heart Disease outcome predictions, where higher values are better. \\

\section{Results}
\subsection{Tests of the MCAR and MAR data set assumptions}
Table 5 showcases the results after performing Little's test to ensure each dataset was created in a MAR and MCAR fashion and in varying amounts (5, 10, 30, and 50 \%). As can be seen in the last column of Table 5 the test statistic are significant when MAR data was created as conditional on another column and not significant when values were removed at random. These results are in accordance with the ultimate goal of being able to compare how data being either MAR or MCAR influences imputed methodologies.

\begin{table}[!h]
 \caption{Little's Test Results for Datasets}
  \centering
  \begin{tabular}{lllll}
    \toprule
    Dataset & Missing type &\% missing & Num. instances missing & test stat,  (p-value)\\
    \midrule
    Diamond & MAR & 5 & 2696 & 10,693 (0.000)\\
    Diamond & MAR & 10 & 5392 & 18,065.77 (0.000)\\
    Diamond & MAR & 30 & 16176 & 41,009.07, (0.000)\\
    Diamond & MAR & 50 & 26960 & 41,409.31, (0.000)\\
    Diamond & MCAR & 5 & 2696 & 1.899, (0.984)\\
    Diamond & MCAR & 10 & 5392 & 6.222, (0.622)\\
    Diamond & MCAR & 30 & 16176 & 7.472, (0.487)\\
    Diamond & MCAR & 50 & 26960& 5.258, (0.730)\\
    \midrule
    Housing & MAR & 5 & 535 & 964.022, (0.000)\\
    Housing & MAR & 10 & 1069 & 2017.706, (0.000)\\
    Housing & MAR & 30 & 3208 & 5451.812, (0.000)\\
    Housing & MAR & 50 & 5346 & 7284.074, (0.000)\\
    Housing & MCAR & 5 & 535 & 1.307, (0.934)\\
    Housing & MCAR & 10 & 1069 & 1.087, (0.955)\\
    Housing & MCAR & 30 & 3208 & 1.867, (0.867)\\
    Housing & MCAR & 50 & 5346 & 1.833, (0.872)\\
    \midrule
    Heart Disease & MAR & 5 & 2389 & 14376.354, (0.000)\\
    Heart Disease & MAR & 10 & 4779 & 22067.449, (0.000)\\
    Heart Disease & MAR & 30 & 14336 & 31409.706, (0.000)\\
    Heart Disease & MAR & 50 & 23893 & 30324.046, (0.000)\\
    Heart Disease & MCAR & 5 & 2389 & 16.019, (0.099)\\
    Heart Disease & MCAR & 10 & 4779 & 14.994, (0.132)\\
    Heart Disease & MCAR & 30 & 14336 & 15.230, (0.124)\\
    Heart Disease & MCAR & 50 & 23893 & 5.566, (0 .850)\\
    \bottomrule
  \end{tabular}
  \label{tab:table}
\end{table}

\subsection{Testing Imputed Values Accuracy (F1)}
Tables 6, 7, and 8 show the F1 values across imputed missing cells. 

\begin{table}[!h]
 \caption{F1 values following imputation of diamond dataset, stratified by type and amount missing}
  \centering
  \begin{tabular}{llllll}
    \toprule
    Missing Type    & Num. missing   & KNN & MICE$^1$ $\pm$ Std. err & Datawig$^1$ $\pm$ Std. err & IRT\\
    \hline
    \hline
    MAR & 2,696 & 0.192 & 0.161 $\pm$ 0.003 & 0.192 $\pm$ 0.007 & 0.289\\
    MAR & 5,392 & 0.201 & 0.159 $\pm$ 0.001 & 0.207 $\pm$ 0.002 & 0.238\\
    MAR & 16,176 & 0.199 & 0.162 $\pm$ 0.001 & 0.202 $\pm$ 0.003 & 0.166\\
    MAR & 26,960 & 0.185 &  0.153 $\pm$ 0.0005 & 0.191 $\pm$ 0.008 & 0.185\\
    \hline
    \multicolumn{2}{c@{}}{Marginal means} &  0.194 & 0.159 & 0.198 & 0.212\\
    \hline
    MCAR & 2,696 & 0.231 & 0.164 $\pm$ 0.003 & 0.222 $\pm$ 0.002 &  0.208\\
    MCAR & 5,392 & 0.223 & 0.158 $\pm$ 0.002 & 0.218 $\pm$ 0.001 & 0.213 \\
    MCAR & 16,176 & 0.219 & 0.159 $\pm$ 0.001 & 0.217 $\pm$ 0.002 & 0.212\\
    MCAR & 26,960 & 0.219 & 0.157 $\pm$ 0.001 & 0.221 $\pm$ 0.001 &  0.209\\
    \hline
    \multicolumn{2}{c@{}}{Marginal means} &  0.223 & 0.160 & 0.219 & 0.211\\
    \bottomrule
  \end{tabular}
  \footnotetext[1]{both MICE and Datawig have inherent randomness as part of the imputation algorithms, and thus repeated imputed datasets (5 for each) and their standard errors have been created for these methodologies throughout the text}
  \label{tab:table}
\end{table}

\begin{table}[!h]
 \caption{F1 values following imputation of housing dataset, stratified by type and amount missing}
  \centering
  \begin{tabular}{llllll}
    \toprule
    Missing Type    & Num. missing   & KNN & MICE$^1$ $\pm$ Std. err  & Datawig$^1$ $\pm$ Std. err  & IRT\\
    \hline
    MAR & 535 & 0.328 & 0.203 $\pm$ 0.008 & 0.523 $\pm$ 0.0004 & 0.523 \\
    MAR & 1069 & 0.332 & 0.206 $\pm$ 0.008 & 0.526 $\pm$ 0.0002 & 0.527\\
    MAR & 3208 & 0.298 & 0.204 $\pm$ 0.004 & 0.515 $\pm$ 0.005 & 0.522\\
    MAR  & 5346 & 0.187 & 0.198 $\pm$ 0.002 & 0.438 $\pm$ 0.004 & 0.456\\
    \hline
    \multicolumn{2}{c@{}}{Marginal means} &  0.286 & 0.203 & 0.501 & 0.507\\
    \hline
    MCAR & 535 & 0.263 & 0.197 $\pm$ 0.005 & 0.566 $\pm$ 0.003 & 0.563 \\
    MCAR & 1069 & 0.264 & 0.193 $\pm$ 0.005 & 0.545 $\pm$ 0.0004 & 0.544\\
    MCAR  & 3208 & 0.255 & 0.194 $\pm$ 0.003 & 0.549 $\pm$ 0.0008 & 0.550\\
    MCAR & 5346 & 0.258 & 0.194 $\pm$ 0.002 & 0.546 $\pm$ 0.0006 & 0.553\\
    \hline
    \multicolumn{2}{c@{}}{Marginal means} & 0.260 & 0.195  & 0.551 & 0.553\\
    \bottomrule
  \end{tabular}
  \footnotetext[1]{both MICE and Datawig have inherent randomness as part of the imputation algorithms, and thus repeated imputed datasets (5 for each) and their standard errors have been created for these methodologies throughout the text}
  \label{tab:table}
\end{table}

\begin{table}[!h]
 \caption{F1 values following imputation of heart disease dataset, stratified by type and amount missing}
  \centering
  \begin{tabular}{llllll}
    \toprule
    Missing Type    & Num. missing   & KNN & MICE$^1$ $\pm$ Std. err  & Datawig$^1$ $\pm$ Std. err  & IRT\\
    \hline
    MAR & 2,389 & 0.852 & 0.662 $\pm$ 0.005 & 0.866 $\pm$ 0.001 & 0.842\\
    MAR & 4,779 & 0.798 & 0.632 $\pm$ 0.003 & 0.836 $\pm$ 0.0003 & 0.798\\
    MAR & 14,336 & 0.664 & 0.579 $\pm$ 0.002 & 0.753 $\pm$ 0.001 & 0.720\\
    MAR & 23,893 & 0.655 & 0.565 $\pm$ 0.0003 &  0.564 $\pm$ 0.015 & 0.691\\
    \hline
    \multicolumn{2}{c@{}}{Marginal means} & 0.742 & 0.610 & 0.755 & 0.763 \\
    \hline
    MCAR & 2,389 & 0.699 & 0.570 $\pm$ 0.008 & 0.734 $\pm$ 0.001 & 0.709\\
    MCAR & 4,779 & 0.686 & 0.575 $\pm$ 0.003 & 0.732 $\pm$ 0.0008 & 0.719\\
    MCAR & 14,336 & 0.691 & 0.574 $\pm$ 0.002 & 0.729 $\pm$ 0.0005 & 0.716\\
    MCAR & 23,893 & 0.697 & 0.574 $\pm$ 0.001 & 0.729 $\pm$ 0.0003 & 0.713\\
    \hline
    \multicolumn{2}{c@{}}{Marginal means} & 0.693 & 0.726 & 0.730 & 0.714 \\
    \bottomrule
  \end{tabular}
  \footnotetext[1]{both MICE and Datawig have inherent randomness as part of the imputation algorithms, and thus repeated imputed datasets (5 for each) and their standard errors have been created for these methodologies throughout the text}
  \label{tab:table}
\end{table}

In the Diamond dataset, the ordinal variable (5-levels) of 'color' category was imputed. There was a significant main effect of methodologies collapsed across MAR and MCAR data sets (F(3,24)=13.3, p $<$ .001). Follow-up tests showed that KNN, DataWig, and IRT all performed significantly better than MICE in reproducing the missing values. Overall, the F1 value for this data set across all methodologies was 0.20, indicating that this is a difficult imputation task.\\

In the Housing dataset, where the imputed variable was a nominal categorical variable, there was also a main effect of methodologies collapsed across MAR and MCAR data sets (F(3,24)=243.35, p $<$ .001). Follow-up tests showed that MICE performed significantly poorer than KNN, DataWig and IRT; KNN and MICE performed significantly poorer than DataWig and IRT. DataWig and IRT performed similarly. The same pattern across methodologies emerged in the follow-up tests. Overall, the F1 value for this data set across all methodologies was 0.38, indicating that this is not as difficult a task as an ordinal categorical imputation, but is still difficult.\\

In the Heart Disease dataset, where the imputed variable was a binary category, there was a marginal effect of missing type of data collapsed across methodology (F(1,24)=9.38, p$<$.001). Overall, the F1 value for this data set across all methodologies was 0.70, indicating that this imputation task is a relatively easy one. Performing a visual inspection of tables 6-8, increasing the percentage of missing items most prominently impacts the F1 scores when the items are missing at random (MAR).\\

\subsection{Effects on Machine Learning Outcomes}
Gradient Boosting Regressor and XGBoost machine learning algorithms outperformed others tested, the results of which are reported for each dataset in Tables 9-11. At the bottom of each table is the recorded performance in root-mean-square error (RMSE) in regression tasks and AUC in the classification task. In a regression and categorical ordinal model (Table 9), MICE and KNN algorithms minimized RMSE the most in both MCAR and MAR data, while Datawig was performed the poorest in MCAR data and IRT worst in MAR data. During a regression task where the imputed variable was a nominal categorical one (Table 10), IRT minimized RMSE the most in both MCAR and MAR datasets, the benefits of which were most exaggerated in MCAR data. In a classification and binary imputation task (Table 11), nearly all methods performed equally well.\\

\begin{table}[!h]
 \caption{RMSE for Diamond dataset}
  \centering
  \begin{tabular}{llllll}
    \toprule
    Type & Missing (\%)  & KNN & MICE$^1$  $\pm$ Std. err.& Datawig$^1$ $\pm$ Std. err.& IRT\\
    \midrule
    MAR & 2,696 (5) & 0.224 &  0.224 $\pm$ 0.00001 & 0.216 $\pm$ 0.00005 & 0.280\\
    MAR & 5,392 (10) & 0.223 & 0.224 $\pm$ 0.00002 & 0.216 $\pm$ 0.00003 & 0.280\\
    MAR & 16,176 (30) & 0.223 & 0.225 $\pm$ 0.00002 & 0.215 $\pm$ 0.0002 & 0.278\\
    MAR & 26,960 (50)& 0.222 & 0.227 $\pm$ 0.00001 & 0.214 $\pm$ 0.0002 & 0.267\\
    \hline
    MCAR & 2,696 (5) & 0.230 & 0.232 $\pm$ 0.0001 & 0.373 $\pm$ 0.0004 & 0.283\\
    MCAR & 5,392 (10)& 0.234 & 0.238 $\pm$ 0.0002 & 0.473 $\pm$ 0.0004 & 0.283\\
    MCAR & 16,176 (30) & 0.249 & 0.258 $\pm$ 0.0002 & 0.704 $\pm$ 0.0006 & 0.285\\
    MCAR & 26,960 (50) & 0.262 & 0.270 $\pm$ 0.0002 & 0.836 $\pm$ 0.0015 & 0.286\\
    \midrule
    & Original data & RMSE: & 0.2241\\
    \bottomrule
  \end{tabular}
  \label{tab:table}
  \footnotetext[1]{both MICE and Datawig have inherent randomness as part of the imputation algorithms, and thus repeated imputed datasets (5 for each) and their standard errors have been created for these methodologies throughout the text}
\end{table}

\begin{table}[!h]
 \caption{RMSE for Housing dataset}
  \centering
  \begin{tabular}{llllll}
    \toprule
    Type & Missing (\%)   & KNN & MICE$^1$  $\pm$ Std. err.& Datawig$^1$ $\pm$ Std. err.& IRT\\
    \midrule
    MAR & 535 & 0.496 & 0.692 $\pm$ 0.0107 & 0.635 $\pm$ 0.0084 & 0.488\\
    MAR & 1,069 & 0.444 & 0.737 $\pm$ 0.0294 & 0.587 $\pm$ 0.0100 & 0.614\\
    MAR & 3,208 & 0.606 & 0.678 $\pm$ 0.0009 & 0.586 $\pm$ 0.0034 & 0.607\\
    MAR  & 5,346 & 0.582 & 0.695 $\pm$ 0.0161 & 0.610 $\pm$ 0.0112 & 0.641\\
    \hline
    MCAR & 535 & 0.760 & 0.745 $\pm$ 0.0023 & 0.591 $\pm$ 0.0101 & 0.445\\
    MCAR & 1,069 & 0.752 & 0.749 $\pm$ 0.0026 & 0.670 $\pm$ 0.0037 & 0.645\\
    MCAR  & 3,208 & 0.796 & 0.768 $\pm$  0.0082 & 0.625 $\pm$ 0.0072 & 0.455\\
    MCAR & 5,346 & 0.809 &  0.806 $\pm$  0.0059 &  0.765 $\pm$ 0.0048& 0.554\\
    \midrule
    & Original data & RMSE: & 0.4380\\
    \bottomrule
  \end{tabular}
  \label{tab:table}
  \footnotetext[1]{both MICE and Datawig have inherent randomness as part of the imputation algorithms, and thus repeated imputed datasets (5 for each) and their standard errors have been created for these methodologies throughout the text}
\end{table}

\begin{table}[!h]
 \caption{AUC for Heart Disease Dataset}
  \centering
  \begin{tabular}{llllll}
    \toprule
    Type & Missing (\%)   & KNN & MICE$^1$  $\pm$ Std. err.& Datawig$^1$ $\pm$ Std. err.& IRT\\
    \midrule
    MAR & 2,389 & 0.832 & 0.831 $\pm$ 0.0001 & 0.831 $\pm$ 0.0001 & 0.830\\
    MAR & 4,779 & 0.830 & 0.830 $\pm$ 0.0002 & 0.831 $\pm$ 0.0002 & 0.832 \\
    MAR & 14,336 & 0.827 & 0.828 $\pm$ 0.0001 & 0.829 $\pm$ 0.0003 & 0.827\\
    MAR & 23,893 & 0.827 & 0.827 $\pm$ 0.0003 & 0.828 $\pm$ 0.0002 & 0.829\\
    \hline
    MCAR & 2,389 &  0.829 & 0.831 $\pm$ 0.0003 & 0.832 $\pm$ 0.0004 & 0.832\\
    MCAR & 4,779 & 0.831 & 0.830 $\pm$ 0.0005 & 0.831 $\pm$ 0.0002 & 0.831\\
    MCAR & 14,336 & 0.829 & 0.828 $\pm$ 0.0005 & 0.829 $\pm$ 0.0002 & 0.828\\
    MCAR & 23,893 & 0.828 & 0.826 $\pm$ 0.0003 & 0.828 $\pm$ 0.0002 & 0.829\\
    \midrule
    & Original data AUC: & 0.8313\\
    \bottomrule
  \end{tabular}
  \label{tab:table}
  \footnotetext[1]{both MICE and Datawig have inherent randomness as part of the imputation algorithms, and thus repeated imputed datasets (5 for each) and their standard errors have been created for these methodologies throughout the text}
\end{table}

\subsection{Effects on Machine Learning Outcomes}

Gradient Boosting Regressor and XGBoost machine learning algorithms outperformed others tested and were used in the regression and classification analyses, respectively. The results are reported for each dataset in Tables 9-11. At the bottom of each table is the recorded performance of the full (no missing) data sets. For the Diamond data set, there were significant main effects of imputation methodology collapsed across MAR and MCAR data sets (F(3,24)=9.19, p$<$.001) (DataWig was significantly worse than all other methodologies), missing data type collapsed across methodology (F(1,24)=16.68, p$<$.001) (MAR was better than MCAR, 0.23 versus 0.34, respectively), and an interaction effect was also significant (F(3,24)=11.69, p$<$.001). Post-hoc tests of the interaction showed that IRT was worse than the other methodologies in the MAR condition, and that DataWig was worse than the other methodologies in the MCAR condition.\\

For the House data set, there were significant main effects of imputation methodology collapsed across MAR and MCAR data sets (F(3,24)=12.40, p $<$.001) (DataWig and IRT were superior to the other methodologies), and an interaction (F(3,24)=9.48, p$<$.001. The interaction showed the that the effect of methodology was with the MCAR data and not the MAR data.\\

There were no effects of methodology or missing data type in the Heart Disease data sets. There was generally no notable consistent change, using a visual inspection of the tables, when increasing the percentage of missing data in these results. The one exception to this pattern was that of the DataWig MCAR for the Diamond data set. These rather large values are much higher than the other imputation methodologies or the original data set value of 0.2241. As the amount of missing data increased, the RMSE values increased quite substantially (0.37, 0.47, 0.71 to 0.84).\\

\section{Discussion}
The results suggest that IRT-based imputation is a viable alternative to some of the more established methods for categorical imputation. It returned more accurate values than DataWig for the Diamond data (ordinal) and more accurate values than KNN and MICE for the Housing data (nominal). \\
In terms of the predictive utility of these substitutions, IRT was superior to DataWig for the Diamond data and superior to KNN and MICE for the Housing data. More nuanced findings indicated that IRT missing value replacement resulted in poorer predictive utility for the Diamond MAR data than other methodologies, but was better than DataWig in predictive utility for the Diamond MCAR data.\\
There were somewhat mixed findings regarding the effect of and underlying structure to the 'missingness' (i.e., whether the data were MCAR versus MAR). For the housing data, MCAR data were more accurately reproduced by MICE, DataWig and IRT than were MAR data. However, in the Heart Disease data, MAR missing data were more accurately reproduced than MCAR missing data for all imputation methodologies. In the Diamond and Housing data sets, there was better predictive utility for the MAR than MCAR data. Thus, on balance, there seems to be a somewhat better result when the substitutions are based on some sort of structure within the data. This makes sense insofar as the methodologies are utilizing other information in the data sets to impute missing values. If the missingness is completely random, then there is very little for the methodologies to capitalize on when converging on a specific value for the missing cells. \\
While the amount of missing data was manipulated, there did not seem to be a very large effect of this variable on the results and was not tested empirically. The one exception was the predictive utility of the Diamond data when MCAR missing data were imputed by DataWig; as the proportion of missing data increased, there was obviously an impediment to the features’ overall predictive utility.\\
One notable finding was that the ordinal categorical data were most difficult for all imputation techniques, followed by the nominal imputations, with the binary imputations most easily addressed. This intriguing finding is quite possibly a result of the one-hot encoding limitation required by algorithms such as KNN and MICE, and distributional effects of ordinal categories. Binary imputation with two distinct classes leaves fewer available options, and thus being correct by chance is higher as a result.\\
There were no effects of imputation on high blood pressure (binary) from the heart disease data (binary missing data), indicating that none were superior/inferior with this type of data, with accuracy or predictive utility. This may be due to blood pressure existing in 2 distinct states. On closer examination of the heart disease where 23,893 blood pressure values were missing (MCAR: 1-13,694, 0-10,209) and (MAR: 1-16,660, 0-7,233), where '1' denote high blood pressure and '0' does not, there exists significant imbalance in the two data sets. However, results were very similar.\\
Although DataWig is often described as being superior to other imputation methods in that it handles different types of data, it did not perform as well as the others in this study on some data sets – more poorly on the ordinal data than all the others, no better than IRT on the nominal data, and no better than any of the others on the binary data. There is also the circularity issue in using DataWig as it uses the outcome variable when estimating missing values. As per DataWig’s documentation, it requires at least 10 times more rows than the unique number of categories to impute missing values for categorical variables. In the current study, it had difficulty imputing a category that appeared infrequently within a categorical variable.\\
Although not shown here, a strength of IRTCI is when continuous feature values have a non-linear relationship to the outcome, or are highly skewed, modifying the variable to be a categorical estimate may be a very useful alternative. For instance, many lab values in healthcare data are associated with poor health outcomes if they are ’out of range’ - abnormally high or abnormally low. Hypo- and hypernatremia are examples of this. These pose a unique challenge for linear imputation methods. Employing IRTCI, cut points could be made that delineate the normal range (135-145 mmol/L) from abnormally high or abnormally low. Missing values could be imputed under GRM or NRM methodology in IRT. In addition, IRTCI methods can be used with supervised or unsupervised data sets.\\
The IRTCI method was tested on multiple data sets, but as with any novel methodology, there are limitations to the work. One was that the data from a single variable was missing; this was done to control the effects. It is possible that if the structure of the missing data was modified, the results might change. This remains an open invitation to researchers in other disciplines to perform, as our attempt was to control as many variables as possible to ensure the internal validity of the findings. Second, IRTCI requires the movement between two different software platforms, and moving between them can be a deterrent. Lastly, IRTCI is useful primarily for categorical imputations (binary, nominal, and ordinal) as demonstrated in this study. Another opportunity for future research into this method include adapting it for use with continuous features. IRT protocols allow for categorization of continuous data into many ordinal-level groups (e.g., 10-15). Such a set of would be a ’near continuous’ approximation of the data. Imputed missing values could be mapped back to the distribution from whence it came, allowing for a point estimate of the data. Such an approach would require large data sets to ensure adequate numbers of cases in each group. Additional future work is warranted to demonstrate how this method would perform.\\

\section{Conclusion}
Our findings showcased a novel categorical imputation method - Item Response Theory for Categorical Imputation (IRTCI). Categorical imputation poses some unique problems, unlike multiple imputation based on for continuous, normally distributed data, categorical multiple imputations with many variables results in large numbers of higher order interactions \cite{finch2010imputation}. Most imputation methods used in machine learning require transformation to one-hot encoded values and do not have native methods for handling nominal categories. The IRTIC technique uses a theoretically justified probabilistic approach to imputing the most likely value for a categorical variable. As it is outlined in this study IRTIC presents a viable alternative to existing methods.

\section*{Acknowledgments}
This work was supported by grants U01TR003528 and R01LM013337.

\section*{Conflicts of Interest}
All authors declare that they have no conflicts of interest.

\bibliography{sn-bibliography}

\end{document}